\documentclass[runningheads]{llncs}

\usepackage{graphicx}
\usepackage{comment}
\usepackage{amsmath,amssymb,amsfonts} 
\usepackage{color}

\usepackage{tikz}
\usepackage{cite}
\usepackage{algorithmic}
\usepackage{graphicx}
\usepackage{caption}
\usepackage{subcaption}
\usepackage{textcomp}
\usepackage{xcolor}
\usepackage{multirow}

\newcommand{\ie}{\textit{i.e.}}

\begin{document}
\pagestyle{headings}
\mainmatter
\def\ECCVSubNumber{01}  

\title{Probabilistic Object Classification\\ using CNN ML-MAP layers} 

\titlerunning{Probabilistic object classification}
%
\author{G. Melotti \inst{1} \and
C. Premebida\inst{1}
\and
J. J. Bird\inst{2} \and D. R. Faria\inst{2}
\and
N. Gon\c{c}alves\inst{1} 
}
\authorrunning{G.Melotti, C.Premebida, et al.}
%
\institute{ISR-UC, Universiy of Coimbra,  Portugal.\\ \email{\{gledson.melotti,cpremebida,nunogon\}@isr.uc.pt}\\
\url{https://www.isr.uc.pt} 
\and
ARVIS Lab, Aston University, UK.\\ \email{\{birdj1,d.faria\}@aston.ac.uk}\\
\url{http://arvis-lab.io} 
}

\maketitle

\begin{abstract}
Deep networks are currently the state-of-the-art for sensory perception in autonomous driving and robotics. However, deep models often generate overconfident predictions precluding proper probabilistic interpretation which we argue is due to the nature of the SoftMax layer. To reduce the overconfidence without compromising the classification performance, we introduce a CNN probabilistic approach based on distributions calculated in the network's Logit layer. The approach enables Bayesian inference by means of ML and MAP layers. Experiments with calibrated and the proposed prediction layers are carried out on object classification using data from the KITTI database. Results are reported for camera ($RGB$) and LiDAR (range-view) modalities, where the new approach shows promising performance compared to SoftMax.

\end{abstract}

\section{Introduction}
\label{sec:Intro}

In state-of-the-art research, the majority of CNN-based classifiers (Convolutional neural networks) train to provide normalized prediction-scores of observations given the set of classes, that is, in the interval $[0, 1]$ \cite{Su_2018_ECCV}. Normalized outputs aim to guarantee ``probabilistic" interpretation. However, how reliable are these predictions in terms of probabilistic interpretation? Also, given an example of a non-trained class, how confident is the model? These are the key questions to be addressed in this work. 

Currently, possible answers to these open issues are related to calibration techniques and penalizing overconfident output distributions~\cite{NIPS2019_calib,oncalibration,GabrielPereyra}. Regularization is often used to reduce overconfidence, and consequently overfitting, such as the confidence penalty~\cite{GabrielPereyra} which is added directly to the cost function. Examples of transformation of network weights include $L1$ and $L2$ regularization~\cite{AndrewY}, Dropout~\cite{rivastava14}, Multi-sample Dropout~\cite{MultiSampleDropout} and Batch Normalization~\cite{BatchNormalization}. Alternatively, highly confident predictions can often be mitigated by calibration techniques such as Isotonic Regression~\cite{isotonicregression} which combines binary probability estimates of multiple classes, thus jointly optimizing the bin boundary and bin predictions; Platt Scaling~\cite{plattscaling} which uses classifier predictions as features for a logistic regression model; Beta Calibration~\cite{betacalibration} which is the use of a parametric formulation that considers the Beta probability density function (pdf); and temperature scaling (TS)~\cite{temperaturescaling} which multiplies all values of the Logit vector by a scalar parameter $\frac{1}{T}>0$, for all classes. The value of $T$ is obtained by minimizing the negative log likelihood on the validation set.

\begin{figure}[!t]
     \centering
     \begin{subfigure}[b]{0.3\textwidth}
         \centering
         \includegraphics[width=\linewidth]{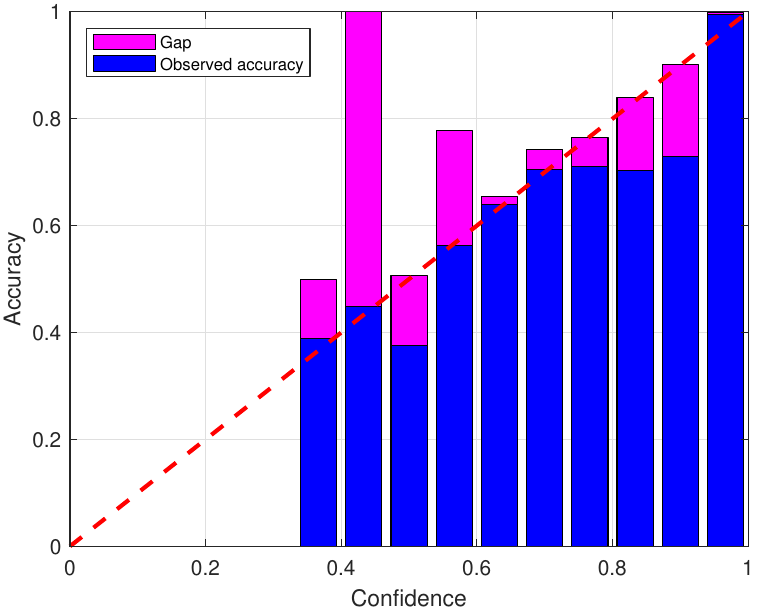}
         \caption{$UC$-$RGB$}
         \label{RD_RGB_NC}
     \end{subfigure}
   \hfill
      \begin{subfigure}[b]{0.3\textwidth}
          \centering
          \includegraphics[width=\linewidth]{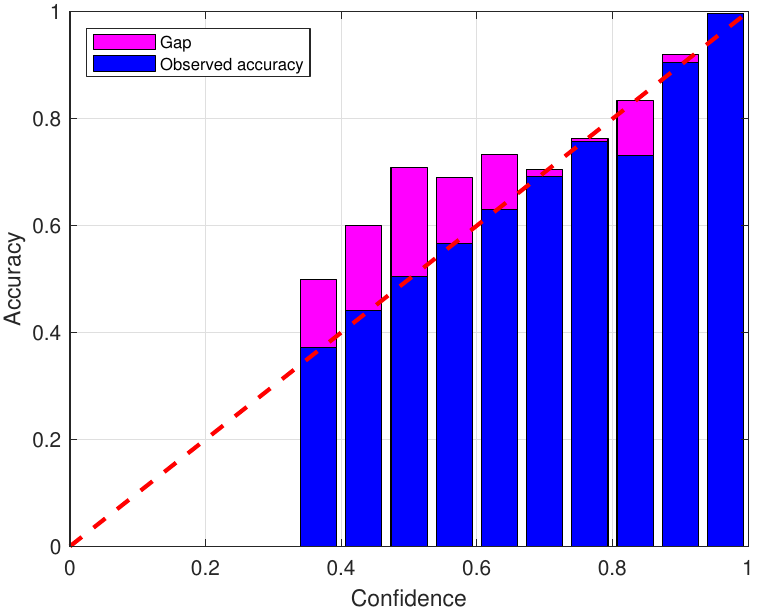}
          \caption{$TS$-$RGB$}
          \label{RD_RGB_TS}
      \end{subfigure}
      \hfill
      \begin{subfigure}[b]{0.3\textwidth}
          \centering
          \includegraphics[width=\linewidth]{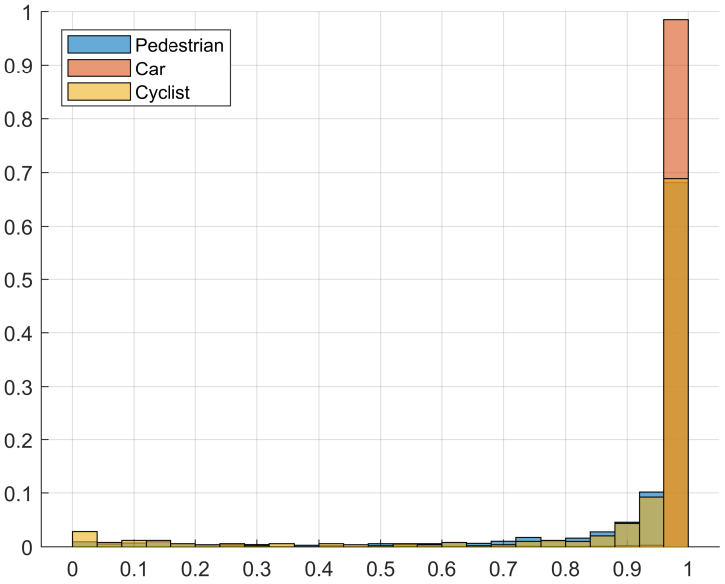}
          \caption{$TS$-$RGB$ SM scores}
          \label{HG_RGB_TS}
      \end{subfigure}
      \\
    \begin{subfigure}[b]{0.3\textwidth}
         \centering
         \includegraphics[width=\linewidth]{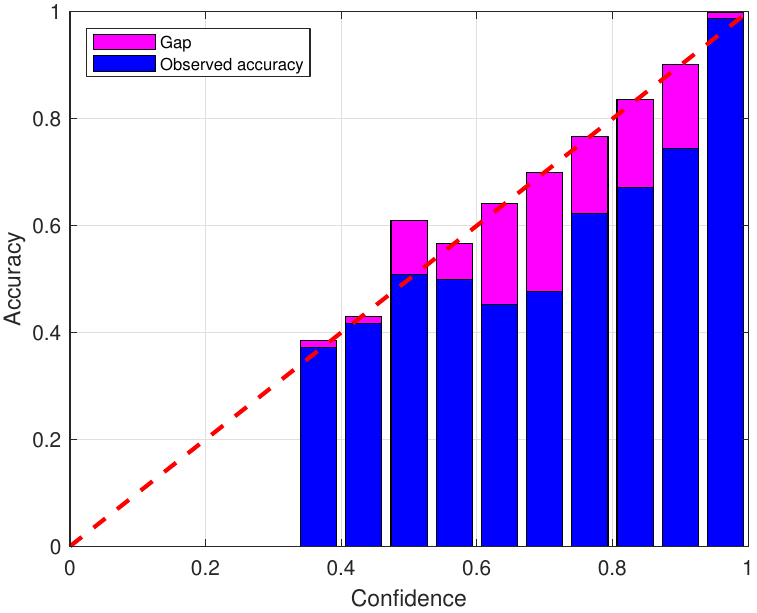}
         \caption{$UC$-$RV$}
         \label{RD_DM_NC}
     \end{subfigure}
   \hfill
      \begin{subfigure}[b]{0.3\textwidth}
          \centering
          \includegraphics[width=\linewidth]{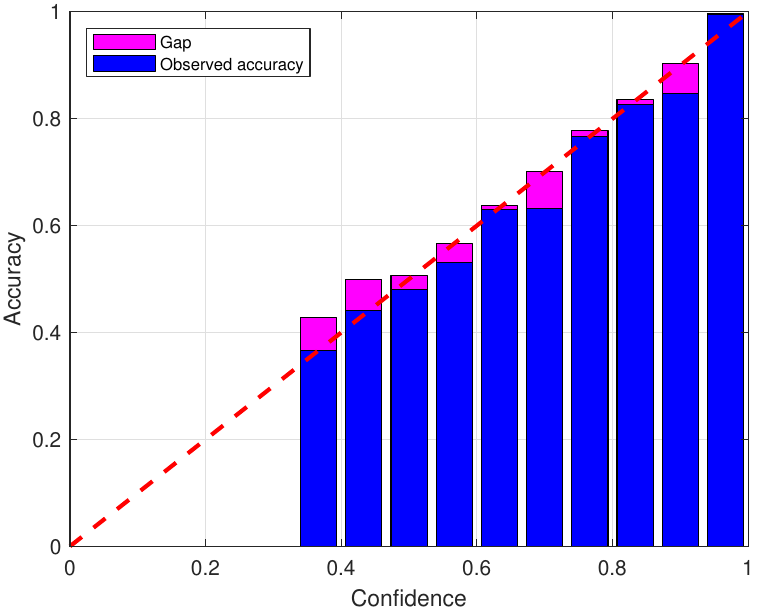}
          \caption{$TS$-$RV$}
          \label{RD_DM_TS}
      \end{subfigure}
      \hfill
      \begin{subfigure}[b]{0.3\textwidth}
          \centering
          \includegraphics[width=\linewidth]{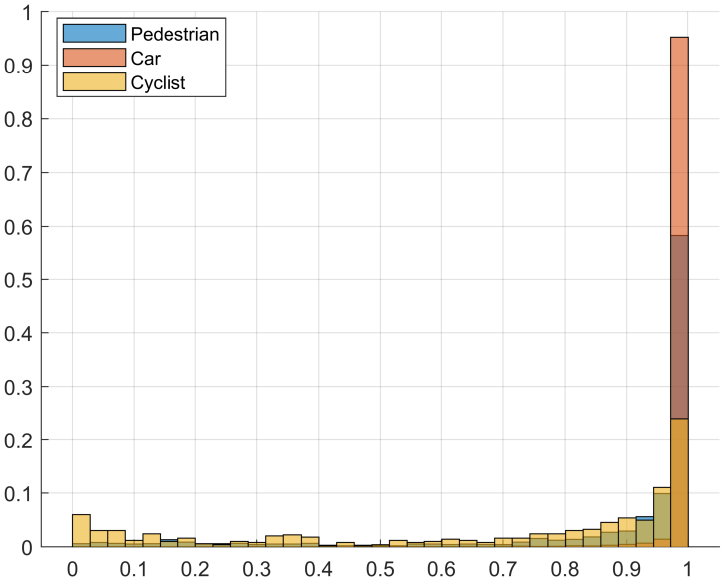}
          \caption{$TS$-$RV$ SM scores}
          \label{HG_DM_TS}
      \end{subfigure}
    \caption{RGB modality reliability diagrams (1$^{st}$ row), on the testing set, for uncalibrated (UC) in (a), and for temperature scaling (TS) in (b), with $T=1.50$. Subfigure (c) shows the distribution of the calibrated prediction scores using SoftMax (SM). The 2$^{nd}$ shows the LiDAR (range-view: RV) modality reliability diagrams in (d) and (e), with $T=1.71$, while in (f) is the prediction-score distribution. Note that (c) and (f) are still overconfident post-calibration.}
    \label{RD_RGB_HG_TS}
\end{figure}

Typically, post-calibration predictions are analysed via reliability diagram representations~\cite{beytscal,oncalibration}, which illustrate the relationship the of the model's prediction scores in regards to the true correctness likelihood~\cite{Niculescu}. Reliability diagrams show the expected accuracy of the examples as a function of confidence \ie, the maximum SoftMax value. The diagram illustrates the identity function should it be perfectly calibrated, while any deviation from a perfect diagonal represents a calibration error~\cite{beytscal,oncalibration}, as shown in Fig. \ref{RD_RGB_NC} and \ref{RD_RGB_TS} with the uncalibrated ($UC$) and temperature scaling ($TS$) predictions on the testing set. Otherwise, Fig. \ref{HG_RGB_TS} shows the distribution of scores (histogram), which is, even after $TS$ calibration, still overconfident. Consequently, calibration does not guarantee a good balance of the prediction scores and may jeopardize adequate probability interpretation.

 \noindent

Complex networks such as multilayer perceptron (MLPs) and CNNs are generally overconfident in the prediction phase, particularly when using the baseline SoftMax as the prediction function, generating ill-distributed outputs \ie, values very close to zero or one~\cite{oncalibration}. Taking into account the importance of having models grounded on proper probability assumptions to enable adequate interpretation of the outputs, and then making reliable decisions, this paper aims to contribute to the advances of multi sensor ($RGB$ and LiDAR) perception for autonomous vehicle systems \cite{MartinICCV,ga,melotti_icarsc} by using pdfs (calculated on the training data) to model the Logit-layer scores. Then, the SoftMax is replaced by a Maximum Likelihood ($ML$), or by a Maximum A Posteriori ($MAP$), as prediction layers, which provide a smoother distribution of predictive values. Note that it is not necessary to re-train the CNN \ie, this proposed technique is practical.

\begin{figure}[!t]
     \centering
     \begin{subfigure}[b]{0.45\textwidth}
         \centering
         \includegraphics[width=\textwidth]{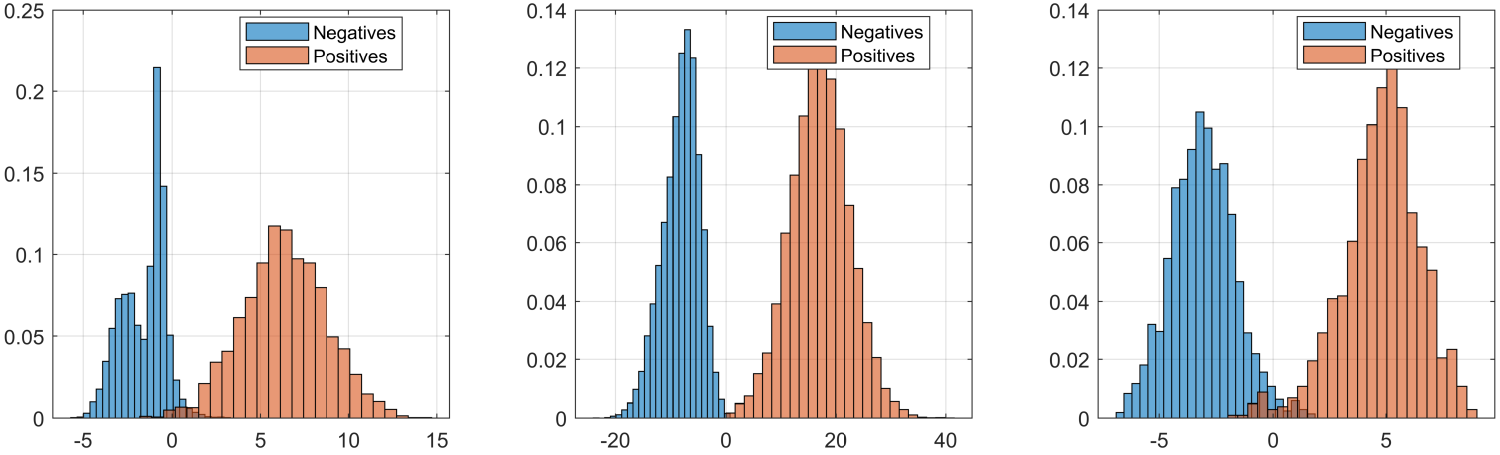}
         \caption{Logit-layer scores, $RGB$.}
         \label{HG_RGB}
     \end{subfigure}
      \hfill
      \begin{subfigure}[b]{0.45\textwidth}
          \centering
          \includegraphics[width=\textwidth]{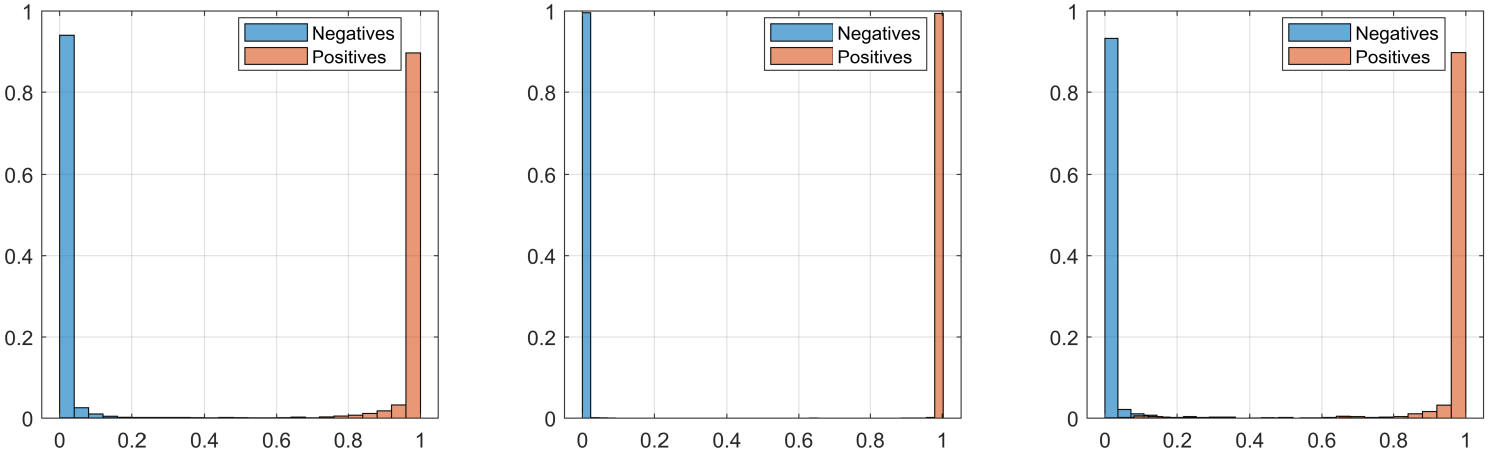}
          \caption{SoftMax-layer scores, $RGB$.}
          \label{SoftMax_RGB}
      \end{subfigure}
      \\
      \begin{subfigure}[b]{0.45\textwidth}
         \centering
         \includegraphics[width=\textwidth]{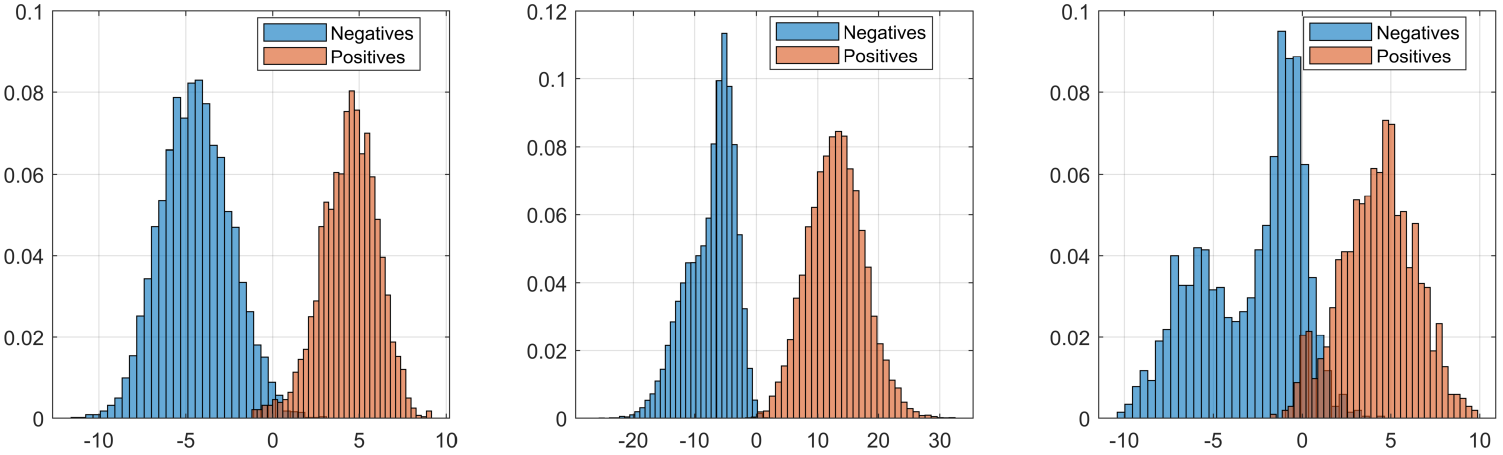}
         \caption{Logit-layer scores, $RV$.}
         \label{HG_DM}
     \end{subfigure}
      \hfill
      \begin{subfigure}[b]{0.45\textwidth}
          \centering
          \includegraphics[width=\textwidth]{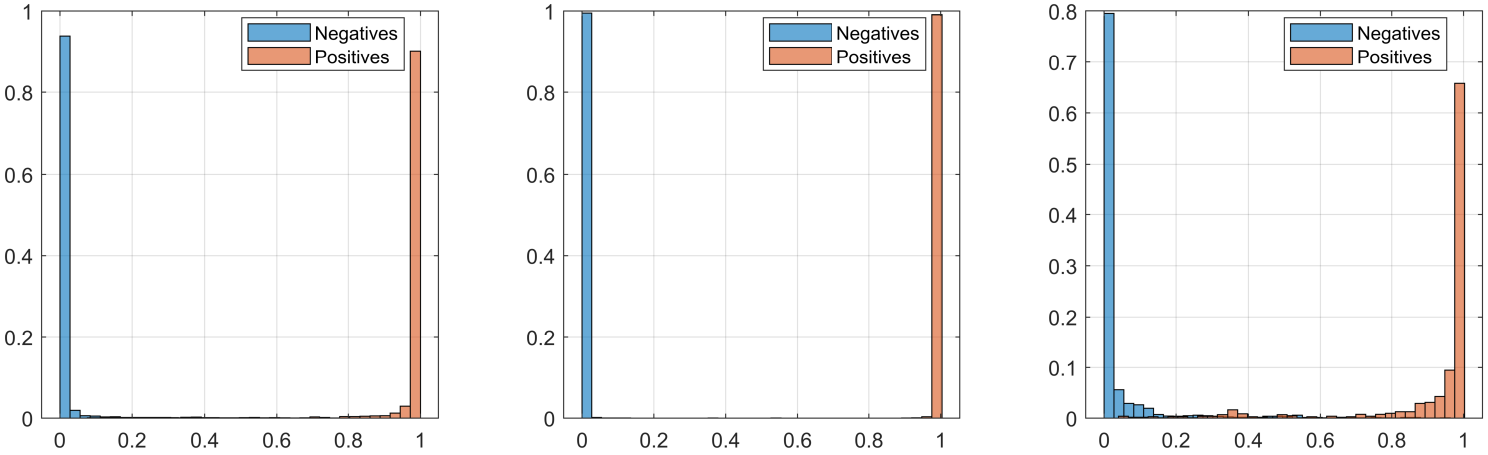}
          \caption{SoftMax-layer scores, $RV$.}
          \label{SoftMax_DM}
      \end{subfigure}
    \caption{Probability density functions (pdf), here modeled by histograms, calculated for the Logit-layer scores for $RGB$ (a) and $RV$ (c) modalities. The graphs in (a,b,c,d) are organized from left-right according to the examples on the Training set (where the positives are in orange). The distribution of the SoftMax prediction-scores in (b) and (d) are an evidence of high confidence.}
    \label{fig:PDFs}
\end{figure} \noindent
\section{Effects of Replacing the SoftMax Layer by a Probability Layer}
\label{sec:method}
The key contribution of this work is to replace the SoftMax-layer (which is a ``hard'' normalization function) by a probabilistic layer (a $ML$ or a $MAP$ layer) during the testing phase. The new layers make inference based on pdfs calculated on the Logit prediction scores using the training set. It is known that the SoftMax scores are overconfident (very close to zero or one), on the other hand the distribution of the scores at the Logit-layer is far-more appropriate to represent a pdf (as shown in Fig. \ref{fig:PDFs}). Therefore, replacement by $ML$ or $MAP$ layers would be more adequate to perform probabilistic inference in regards to permitting decision-making under uncertainty which is particularly relevant in autonomous driving and robotic perception systems.

Let $X_{i}^{L}$ be the output score vector\footnote{The dimensionality of $X$ is proportional to the number of classes.} of the CNN in the Logit-layer for the example $i$, $C_i$ is the target class, and $P(X_{i}^L|C_i)$ is the class-conditional probability to be modelled in order to make probabilistic predictions. In this paper, a non-parametric pdf estimation, using histograms with $25$ (for the $RGB$ case) and $35$ bins (for the $RV$ model), was applied over the predicted scores of the Logit-layer, on the training set, to estimate $P(X^L|C)$. Assuming the priors are uniform and identically distributed for the set of classes $C$, thus a $ML$ is straightforwardly calculated normalizing $P(X_{i}^L|C_i)$, by the $P(X_{i})$ during the prediction phase. Additionally, to avoid `zero' probabilities and to incorporate some uncertainty level on the final prediction, we apply additive smoothing (with a factor equal to $0.01$) before the calculation of the posteriors. Alternatively, a $MAP$ layer can be used by considering, for instance, the \textit{a-priori} as modelled by a Gaussian distribution, thus the $i^{th}$ posterior becomes $P(C_i|X_{i}^L) = P(X_{i}^L|C_i)P(C_i)/P(X_i)$, where $P(C_i) \sim \mathcal{N}(\mu,\,\sigma^{2}) $ with mean $\mu$ and variance $\sigma^{2}$ calculated, per class, from the training set. To simplify, the rationale of using Normal-distributed priors is that, contrary to histograms or more tailored distribution, the Normal pdf fits the data more smoothly.

\section{Evaluation and Discussion}
\label{sec:experiments}

\begin{figure}[!t]
     \centering
     \begin{subfigure}[b]{0.45\textwidth}
         \centering
         \includegraphics[width=\textwidth]{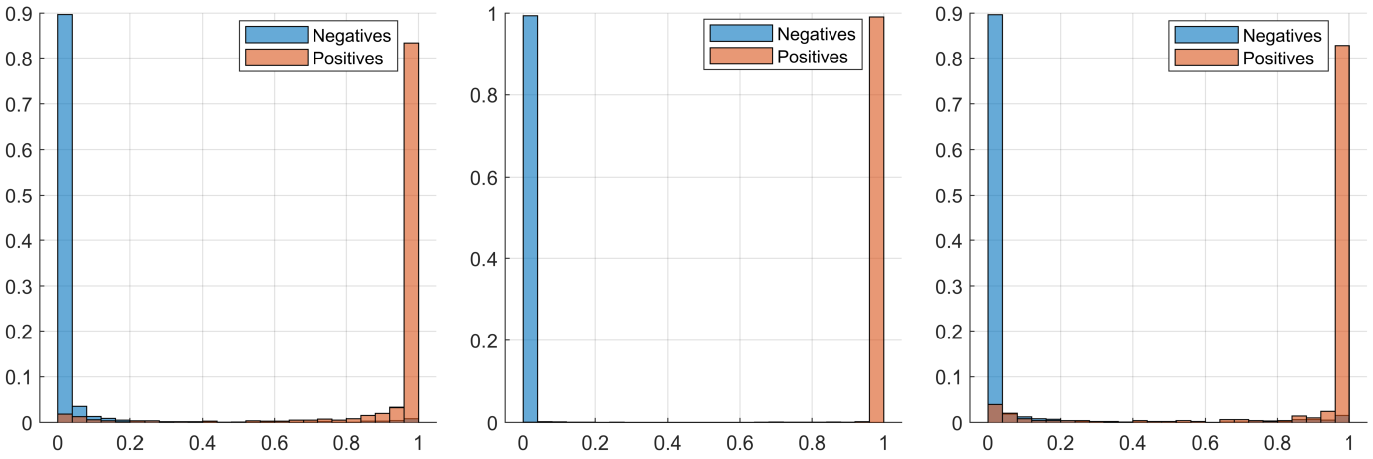}
         \caption{SoftMax-layer scores: $RGB$.}
         \label{PDF_SoftMax_RGB}
     \end{subfigure}
      \hfill
      \begin{subfigure}[b]{0.45\textwidth}
         \centering
         \includegraphics[width=\textwidth]{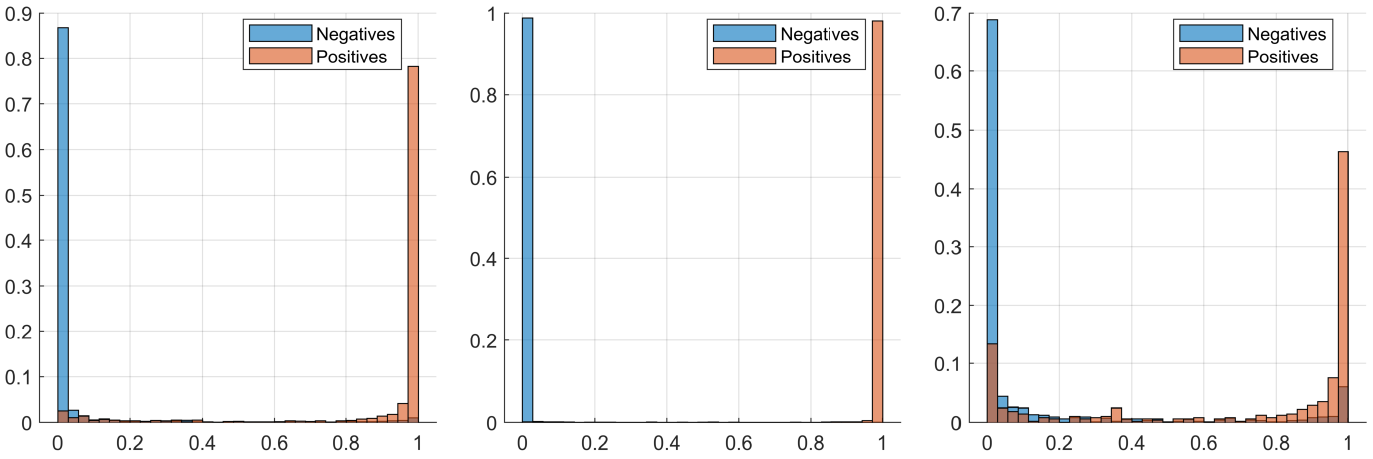}
         \caption{SoftMax-layer scores: $RV$.}
         \label{PDF_SoftMax_DM}
     \end{subfigure}
     \\
      \begin{subfigure}[b]{0.45\textwidth}
          \centering
          \includegraphics[width=\textwidth]{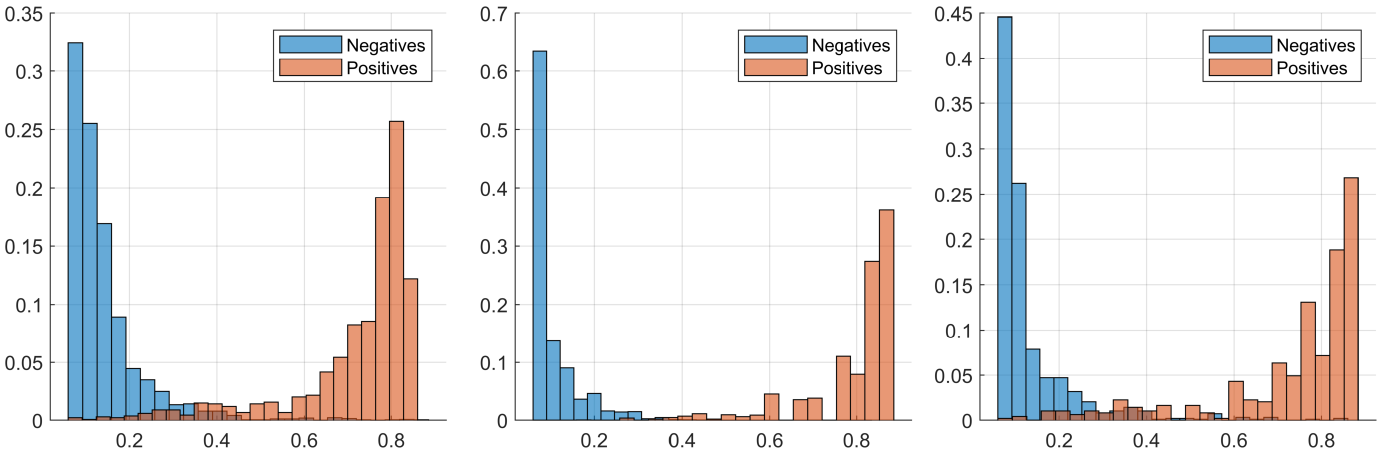}
          \caption{ML-layer scores: $RGB$.}
          \label{PDF_ML_RGB}
      \end{subfigure}
      \hfill
    \begin{subfigure}[b]{0.45\textwidth}
          \centering
          \includegraphics[width=\textwidth]{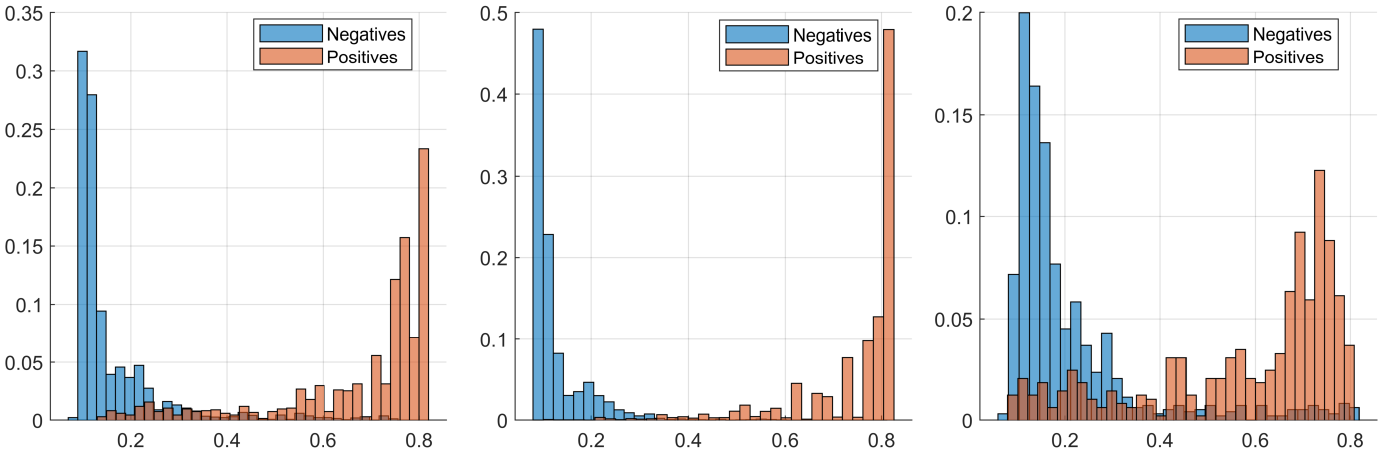}
          \caption{ML-layer scores: $RV$.}
          \label{PDF_ML_DM}
      \end{subfigure}
      \\
       \begin{subfigure}[b]{0.45\textwidth}
         \centering
         \includegraphics[width=\textwidth]{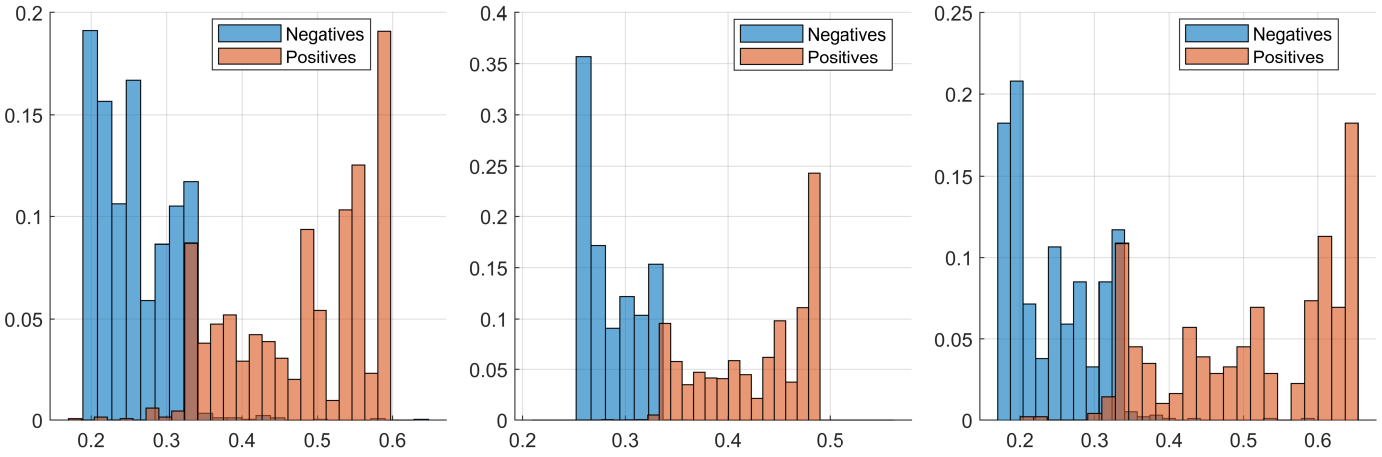}
         \caption{MAP-layer scores: $RGB$.}
         \label{MAP_RGB}
     \end{subfigure}
     \hfill
     \begin{subfigure}[b]{0.45\textwidth}
          \centering
          \includegraphics[width=\textwidth]{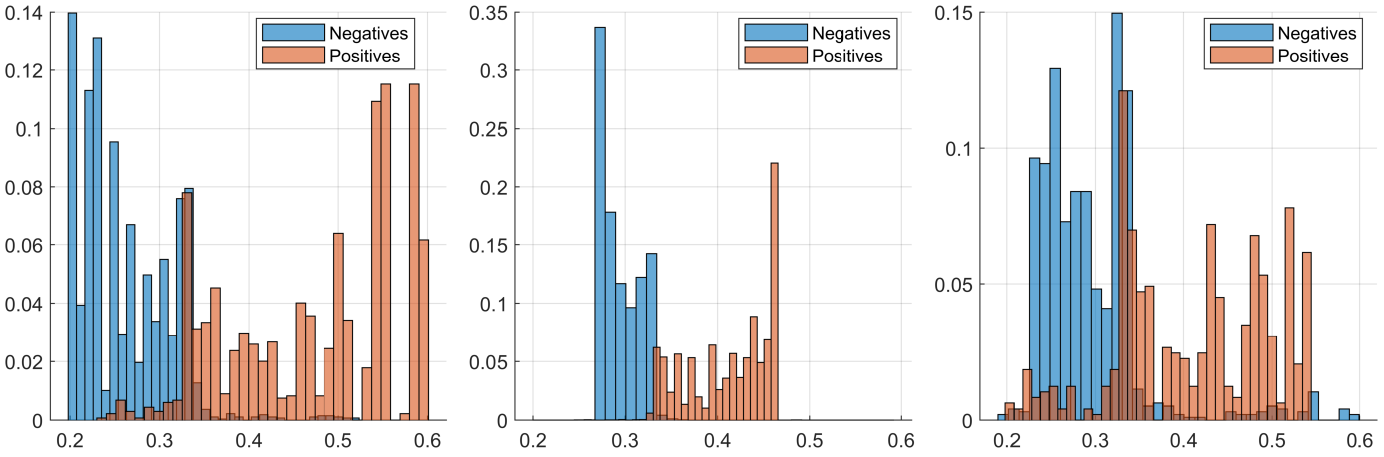}
          \caption{MAP-layer scores: $RV$.}
          \label{MAP_DM}
      \end{subfigure}
    \caption{Prediction scores (\ie, the network outputs), on the Testing set, using SoftMax (baseline solution), ML and MAP layers, for $RGB$ and LiDAR ($RV$) modalities.}
    \label{PDF_RGB}
\end{figure} \noindent

\begin{table}[]
\begin{center}
\caption{Classification performance (\%) in terms of average F-score and $FPR$ for the baseline ($SM$) models compared to the proposed approach of ML and MAP layers. The performance measures on the `unseen' dataset are the average and the variance of the prediction scores.}
\begin{tabular}{c|c c c| c c c}
\hline \hline
Modalities: \,   & \, $SM_{RGB}$\,& \, $ML_{RGB}$ \,& $MAP_{RGB}$ \,&  $SM_{RV}$\, & \, $ML_{RV}$ \,& $MAP_{RV}$  \\ \hline
F-score    \,   & \, $95.89$   \,& \, $94.85$    \,& $95.04$     \, &  $89.48$  \, & \, $88.09$   \,& $87.84$\\ \hline
$FPR$\,   & \, $1.60$    \,& \, $1.19$     \,& $1.14$      \,&  $3.05$   \, & \, $2.22$    \,& $2.33$ \\ \hline
$Ave.Scores_{unseen}$  & \, $0.983$ \, & \, $0.708$ \, & \, $0.397$ \, & \, $0.970$ \, & \, $0.692$ \, & \, $0.394$ \\ \hline
$Var.Scores_{unseen}$  & \, $0.005$  \, & \, $0.025$  \, & \, $0.004$  \, & \, $0.010$  \, & \, $0.017$  \, & \, $0.003$
\end{tabular} \noindent
\label{result}
\end{center}
\end{table} \noindent
\raggedbottom

In this work a CNN is modeled by Inception $V3$. The classes of interest are pedestrians, cars, and cyclists; the classification dataset is based on the KITTI $2D$ object~\cite{ga}, and the number of training examples are $\{2827, 18103, 1025\}$ for `ped', `car', and `cyc'. The testing set is comprised of $\{1346, 8620, 488\}$ examples respectively.

\begin{figure}[!t]
\begin{subfigure}[t]{0.49\textwidth}
          \centering
          \includegraphics[width=\textwidth]{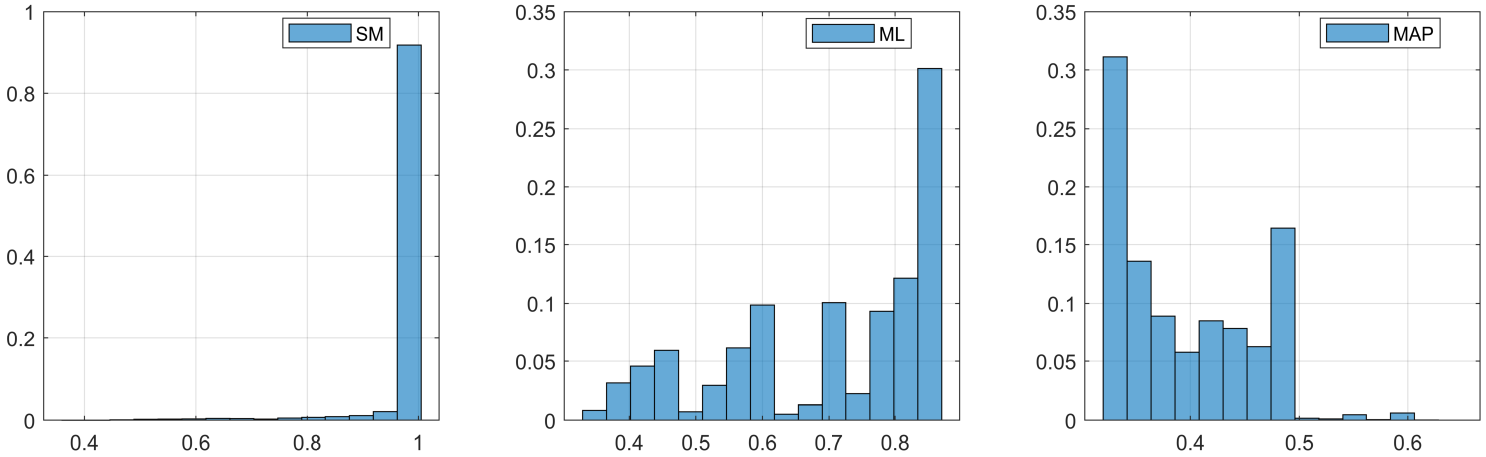}
          \caption{SoftMax (SM), ML, and MAP scores on the $RGB$ unseen set.}
          \label{HG_SoftMax_RGB_US}
      \end{subfigure}
      \hfill
      \begin{subfigure}[t]{0.49\textwidth}
          \centering
          \includegraphics[width=\textwidth]{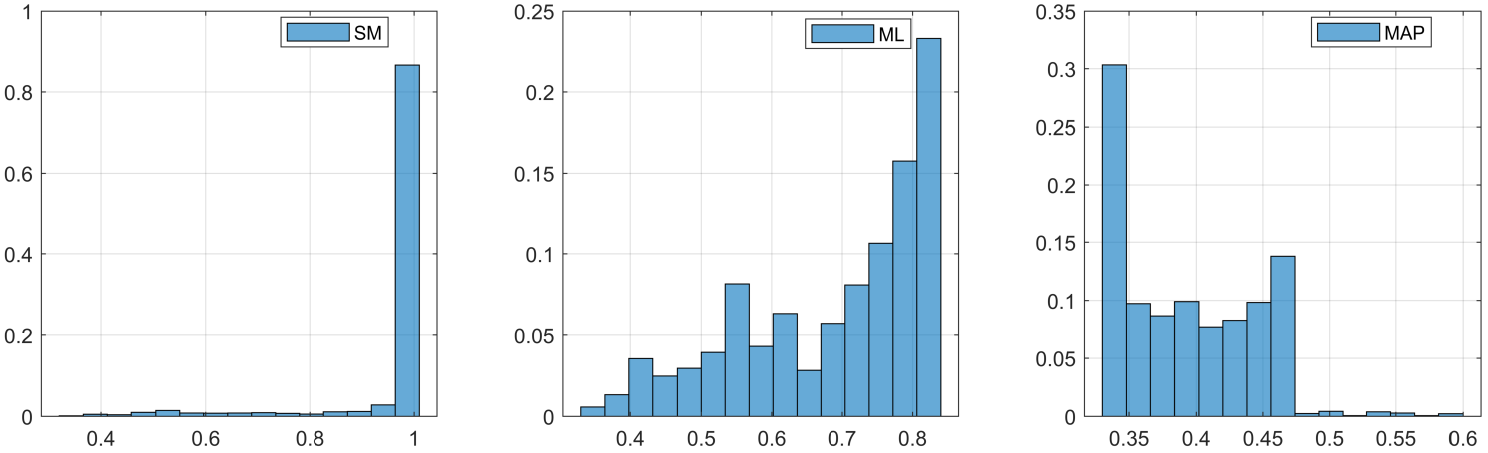}
          \caption{SoftMax (SM), ML, and MAP scores on the $RV$ unseen set.}
          \label{HG_SoftMax_DM_US}
      \end{subfigure}
    \caption{Prediction scores, on the unseen data (comprising non-trained classes: `person\_sit.', `tram', `trees/trunks', `truck', `vans'), for the networks using SoftMax-layer (left-most side), and the proposed ML (center) and MAP (right-side) layers.}
    \label{fig:Unseen}
\end{figure} \noindent

The output scores of the CNN indicate a degree of certainty of the given prediction. The ``certainty level" can be defined as the confidence of the model and, in a classification problem, represents the maximum value within the SoftMax layer \ie, equal to one for the target class. However, the output scores may not always represent a reliable indication of certainty with regards to the target class, especially when unseen or non-trained examples/objects occur in the prediction stage; this is particularly relevant for a real-world application involving autonomous robots and vehicles since unpredictable objects are highly likely to be encountered. With this in mind, in addition to the trained classes (`ped', `car', `cyc'), a set of untrained objects are introduced: `person\_sit.',`tram', `truck', `vans', `trees/trunks' comprised of $\{ 222, 511, 1094, 2914, 45 \}$ examples respectively. All classes with the exception of `trees/trunks' are from the aforementioned KITTI dataset directly, while the former is additionally introduced by this study. The rationale behind this is to evaluate the prediction confidence of the networks on objects that do not belong to any of the trained classes, and thus consequently the consistency of the models can be assessed. Ideally, if the classifiers are perfectly consistent in terms of probability interpretation, the prediction scores would be identical (equal to 1/3) for all the examples on the unseen set on a per-class basis.

Results on the testing set are shown in Table \ref{result} in terms of F-score metric and the average of the FPR prediction scores (classification errors). The average ($Ave.Scores_{unseen}$) and the sample-variance ($Var.Scores_{unseen}$) of the predicted scores are also shown for the unseen testing set. 

To summarise, the proposed probabilistic approach shows promising results since $ML$ and $MAP$ reduce classifier overconfidence, as can be observed in Figures \ref{PDF_ML_RGB}, \ref{PDF_ML_DM}, \ref{MAP_RGB} and \ref{MAP_DM}. In reference to Table \ref{result}, it can be observed that the FPR values are considerably lower than the result presented by a SoftMax (baseline) function. Finally, to assess classifier robustness or the uncertainty of the model when predicting examples of classes untrained by the network, we consider a testing comprised of `new' objects. Overall, the results are exciting since the distribution of the predictions are not extremities as can be observed in Fig. \ref{fig:Unseen}. Quantitatively, the average scores of the network using $ML$ and $MAP$ layers are significantly lower than the SoftMax approach, and thus are less confident on new/unseen negative objects .

%
%
\bibliographystyle{unsrt}
\bibliography{refs}

\end{document}